\PassOptionsToPackage{unicode=true}{hyperref} % options for packages loaded elsewhere
\PassOptionsToPackage{hyphens}{url}
\documentclass[a4paper]{article}
\usepackage{lmodern}
\usepackage{amssymb,amsmath}
\usepackage{ifxetex,ifluatex}
\usepackage{fixltx2e} % provides \textsubscript
\ifnum 0\ifxetex 1\fi\ifluatex 1\fi=0 % if pdftex
  \usepackage[T1]{fontenc}
  \usepackage[utf8]{inputenc}
  \usepackage{textcomp} % provides euro and other symbols
\else % if luatex or xelatex
  \usepackage{unicode-math}
  \defaultfontfeatures{Ligatures=TeX,Scale=MatchLowercase}
\fi
\IfFileExists{upquote.sty}{\usepackage{upquote}}{}
\IfFileExists{microtype.sty}{%
\usepackage[]{microtype}
\UseMicrotypeSet[protrusion]{basicmath} % disable protrusion for tt fonts
}{}
\IfFileExists{parskip.sty}{%
\usepackage{parskip}
}{% else
\setlength{\parindent}{0pt}
\setlength{\parskip}{6pt plus 2pt minus 1pt}
}
\usepackage{hyperref}
\hypersetup{
            pdftitle={Position Paper: Towards Transparent Machine Learning},
            pdfauthor={Dustin Juliano},
            pdfborder={0 0 0},
            breaklinks=true,
            colorlinks=true,
            urlcolor=blue,
            citecolor=blue,
            linkcolor=blue}
\providecommand{\tightlist}{%
  \setlength{\itemsep}{0pt}\setlength{\parskip}{0pt}}
\ifx\paragraph\undefined\else
\let\oldparagraph\paragraph
\renewcommand{\paragraph}[1]{\oldparagraph{#1}\mbox{}}
\fi
\ifx\subparagraph\undefined\else
\let\oldsubparagraph\subparagraph
\renewcommand{\subparagraph}[1]{\oldsubparagraph{#1}\mbox{}}
\fi

\makeatletter
\def\fps@figure{htbp}
\makeatother

\usepackage[style=numeric,backend=biber,hyperref=true,sorting=none]{biblatex}
\title{Position Paper: Towards Transparent Machine Learning}
\author{Dustin Juliano}
\date{Nov.~1, 2019}

\begin{document}
\maketitle
\begin{abstract}
Transparent machine learning is introduced as an alternative form of
machine learning, where both the model and the learning system are
represented in source code form. The goal of this project is to enable
direct human understanding of machine learning models, giving us the
ability to learn, verify, and refine them as programs. If solved, this
technology could represent a best-case scenario for the safety and
security of AI systems going forward.
\end{abstract}

\hypertarget{introduction}{%
\section{1 Introduction}\label{introduction}}

Current machine learning (ML) systems produce models that are difficult
or impossible to understand. This poses clear challenges with security,
safety, and bias in these deployments. Opaque models also make it
difficult to gain insight into the automated decision-making process.

As a result of this, interpretable or explainable ML have become active
areas of research \autocites{samek2017explainable}{liu2017towards}, with
a notable interest from DARPA \autocite{gunning2017explainable}.
However, those approaches are focused on analyzing models that are
inherently resistant to human understanding due to the way in which they
are represented.

Transparent machine learning (TML) is intended to solve these problems
by producing models and data that we can understand. It would do this by
representing and modifying source code representations. This, in turn,
would result in a potentially self-contained executable that could be
deployed directly.

In addition to the source code model, the TML system itself may be
embedded into the output program. This would give it the ability to
continuously update itself. Embedding, however, is not a requirement;
the learner may be suppressed so that it is not emitted with the final
program. This could be for safety purposes or to ensure stability of the
deployment.

A complex program may require auxiliary information, such as labels,
lookup tables, or a database of some kind. Transparent machine learning
systems will need to produce such data to accurately represent the
possible programs in its search space. While the data can and should be
made part of the source code model, large quantities of information
should be stored externally. This should be in a format that is
appropriate to its size and intended usage.

It it crucial that TML systems target commonly known programming
languages and data formats that can be easily understood. Further, the
source code and data it produces in those languages and formats
\emph{must} be of sufficient clarity so that it can be easily understood
and modified by an engineer of reasonable skill. This is a central
tenant that should take priority over all other considerations, even at
cost of model efficiency. Later on, suggestions will be given on how
both efficiency and legibility might be accomplished, without
permanently sacrificing either.

\hypertarget{long-term-objectives}{%
\subsection{1.1 Long-Term Objectives}\label{long-term-objectives}}

The ultimate goal of TML is for it to target one or more dependently
typed programming languages that feature support for rich
specifications. This would necessarily entail concepts from the work
being done on Deep Specification \autocite{appel2017position}, which is
an ambitious project\footnote{See also: deepspec.org.} that seeks to
formally verify the entire development pipeline, end-to-end, from
application to OS, right down to the hardware.

Then there are the long-term quality goals for the generated source code
itself:

\begin{itemize}
\tightlist
\item
  Dense commenting
\item
  Use of high-level abstractions
\item
  Minimization of complexity
\item
  Use of accelerated hardware
\item
  Multiple language targets
\end{itemize}

Fully solving these goals may require complete or partial general AI.
This may put it into the classification of being a potentially
AI-complete or AI-hard problem \autocite{yampolskiy2012ai}.

\hypertarget{short-term-objectives}{%
\subsection{1.2 Short-Term Objectives}\label{short-term-objectives}}

The immediate goals are to:

\begin{enumerate}
\def\labelenumi{\arabic{enumi}.}
\item
  Create a working transparent machine learning algorithm.
\item
  Have that produce legible source code.
\end{enumerate}

The immediate challenge is just getting a working proof-of-concept. An
early strategy would be to find TML systems that equal or rival the best
ML. The rationale for this is that it would increase research interest.
Legibility, however, must eventually become the highest priority,
otherwise it defeats the purpose of the project; incomprehensible source
code is just another type of opaque model.

\hypertarget{motivation}{%
\section{2 Motivation}\label{motivation}}

One practical benefit of TML is that we would unlock the capability to
produce explicit AI implementations. This could potentially help solve
one of the largest technical challenges of AI safety and security, known
as \emph{the control problem} \autocite{bostrom2016superintelligence}.

Verifying that an opaque ML model adheres to its specification is an
impressive research effort
\autocites{dvijotham2018dual}{singh2018fast}{weng2018towards}{seshia2016towards},
but it is not equivalent to having human-readable source code.
Transparent machine learning, in its most advanced form, would not only
give us an explicit program, but it would be \emph{``correct by
construction''}. This would be a result of it having been generated in
programming languages with the most advanced type systems.

By studying the source code produced by TML we could also gain knowledge
about how it implemented its solution. Instead of creating systems that
just give predictions, we could explore the space of programs that
provide these models of intelligence. This could lead to unique insights
into both the problem domain it is addressing and our understanding of
artificial intelligence itself.

Transparent machine learning can be seen as a form of automated
programming. This could greatly accelerate software and hardware
development. And, because TML must understand source code in order to
modify it, this gives it the potential to audit the software and
hardware that we have created. It could be used to find defects, improve
efficiency, and even port projects to modern languages and platforms.
Thus, it is not just something that will only help us going forward, but
could be used to retroactively upgrade existing software.

Having explicit AI implementations will help us understand how to build
better automation. It may even help us find new research directions to
pursue general AI. As will be discussed ahead, these two research
directions will eventually converge as the most significant challenges
of TML are addressed.

\hypertarget{challenges}{%
\section{3 Challenges}\label{challenges}}

The greatest initial challenge for TML will be in making it as effective
and efficient as current ML systems. It may be the case that computing
TML models will require significantly more resources. The author
believes, however, that the resulting TML model, once found, will run
significantly faster than traditional machine learning deployments. The
rationale for this prediction is that TML source models will be capable
of being natively compiled.

\hypertarget{the-labeling-problem}{%
\subsection{3.1 The Labeling Problem}\label{the-labeling-problem}}

After the initial proof-of-concept is realized there will be another
problem, and it will likely hold sway over TML until general AI is
discovered: \emph{the labeling problem}. This is the challenge of how to
create the most legible source code.

It is highly likely that the first TML models will be incomprehensible
to us. An early example may end up looking something like the following:

\begin{verbatim}
    // Begin prototypes
    double _EB29F654701535CC(
        double _0,
        double _1,
        double _2,
        double _3);
    double _A141F416696C035A(
        double _0,
        double _1,
        double _2,
        double _3,
        double _4,
        double _5,
        double _6,
        double _7,
        double _8);
    double _D1773B8053A78241(
        double _0,
        double _1,
        double _2,
        double _4,
        double _5,
        double _6,
        double _7);
\end{verbatim}

That is clearly not a desirable final outcome, but it would be
acceptable for an initial proof-of-concept.

It is important to remember, however, that the labeling problem is not
just about the naming of identifiers. It is also about the patterns that
are used to realize the source code model.

We need to move away from an abstract function space and into the logic
and conventions of computer programs. This means the use of algorithms
and data structures that are ubiquitous in software engineering.

As computer programs are extremely flexible with what they can entail,
the challenge will be in confining program search to patterns that we
would be likely to use if we knew what the TML system knew about the
data.

The labeling problem also includes the issue of commenting. This implies
that the concepts being used by the transparent learner will have to be
represented in natural language.

There really is no upper-bound on the complexity of comments, as this is
equivalent to the TML system describing its ``thoughts'' or intent in a
way that may be missing from the source code itself. This capability
puts this part of the labeling problem well into the range of problems
only solvable by general AI. It would be sufficient, however, that early
comments were simple enough to just help organize the structure of the
source, and, perhaps, make references to particular features or subsets
of training data.

\hypertarget{program-search}{%
\subsection{3.2 Program Search}\label{program-search}}

Program search is fundamentally hard. The space of programs is what TML
must operate over, and to do so it must intelligently iterate over the
well-formed statements of the language generated by some formal grammar.
For detailed information on formal grammars and languages see
\autocites{martin1991introduction}{chomsky1959certain}{naur1975programming}.

How will search be conducted? Will it be directed or random? A
combination of both? What is the measure of fitness? What are the
parameters of the search? These are just some of the questions that will
have to be answered. Should one use algorithmic complexity
\autocites{kolmogorov1965three}{chaitin1966length}, Shannon entropy
\autocite{shannon1948mathematical}, minimum message length
\autocites{wallace1968information}{wallace1999minimum}{dowe2011mml}, or
some other method?

An important early task for program search will likely be in reducing
the search space itself. Opportunities may be found just by ensuring the
generated source code is syntactically and semantically correct for the
target language.

There is work related to this under the name of Universal Search
\autocites{levin1973universal}{solomonoff1984optimum}, which uses
iterative search algorithms that seek solutions to \emph{inversion
problems} by exhaustively enumerating program descriptions for a
universal Turing machine. Informally, for a given function \(\phi\) and
value \(y\), an inversion problem is concerned with finding a value
\(x\) such that \(\phi(x)=y\). More information on inversion problems
can be found in \autocite{solomonoff2002progress} and
\autocite{gagliolo2007}.

\hypertarget{program-generation}{%
\subsection{3.3 Program Generation}\label{program-generation}}

Super-optimization
\autocites{massalin1987superoptimizer}{rn1992eliminating}{cmelik1995shade}{joshi2002denali}{brain2006toast}{bansal2006automatic}{schkufza2013stochastic}
will be of interest to researchers undertaking transparent machine
learning. This is because TML may need to make many millions of program
permutations in order to approach meaningful solutions, and it will
likely need to be able to do that extremely quickly to be practical.

However, the problem is not merely one of speed. Reducing the number of
times the learner has to iterate is of greater benefit than reducing the
cost of generating the next permutation. While probably not realizable
in practice, a perfect TML system would only require a single iteration.
This allows us to think of the number of iterations required to reach a
solution as a kind of cost function.

What is so interesting about super-optimization is that it has already
been demonstrated to be effective. It proves a partial technical result
towards fully realized TML without ever having been designed for this
purpose. The challenge is in scaling it up and utilizing it on the
entire source model.

\hypertarget{interfacing}{%
\subsection{3.4 Interfacing}\label{interfacing}}

Most of the knowledge about the system calls and the various APIs that
are used in programming are expressed in disparate collections of
natural language. Worse, they are often of uneven quality.

How would the TML system know about these interfaces? It is not
sufficient for it to simply generate correct syntax for the target
programming language. It must be able to entail the semantics of these
programming interfaces.

The only plausible answer seems to be that we would have to formally
specify the semantics and interdependencies between all of the APIs and
system calls that the TML system could possibly rely upon.

That is extremely non-trivial, as it implies the use of dependently
typed programming languages or specification languages specially suited
to the task of interface-level interdependencies. Being dependently
typed does not necessarily make a programming language suitable for
productively writing complex specifications. There is early support
being developed to entail complex state with dependent types in Idris
\autocite{brady2016state}, and it provides a clear case for just how
difficult it is to make dependent types as productive as conventional
programming language constructs.

Related to this challenge is the act of specifying a system so that it
is internally consistent according to a certain design. This is
different from the above problem of entailing the state and
interdependencies between calls to a programming interface. Both are a
form of specification, but the former has the challenge of having to
interface with existing data and foreign functions.

It can take teams of individuals to verify a complex system. One
prominent example is Project Everest \autocite{bhargavan2017everest},
which seeks to create a formally verified drop-in replacement for HTTPS.
There have been others, such as CompCert \autocite{leroy2012compcert}
and seL4 \autocite{klein2009sel4}.

The purpose of canvasing these projects is to show how complicated and
costly it is to do formal verification. The author strongly believes
that the focus needs to be on making the tools easier to use. This would
make verification more practical and open up the field to a wider
audience.

Whether or not the details of specifying such complex rules and logic
can be simplified is an open question. But it is one that must be
investigated if we are to have ubiquitous, formally verified software
and hardware.

The problem of interfacing for TML is also complicated by the need to
target multiple languages. Data formats are often used to communicate
complex information between many of these systems, some of which are far
more involved than just defining a new data type. A good example of this
challenge would be in how TML would generate compute shaders and the
relevant data to perform general-purpose GPU computation.

\hypertarget{language-choice}{%
\subsection{3.5 Language Choice}\label{language-choice}}

Should the target language be functional or imperative? Object-oriented
or procedural? These are important questions and there probably is no
one right answer. However, there are two qualities that any TML target
language should have: simplicity and efficiency. In terms of feature
complexity and expressiveness, less may be more when it comes to the
choice of target language.

It could be useful to target a total programming language, though it
need not be functional, as with \autocite{turner2004total}. While this
would help ensure that every program terminates, it does not resolve the
problem entirely. Consider a program that calls the Ackermann function
\autocite{sundblad1971ackermann} with large positive integer values for
each of its arguments. This would pass a termination check, but the
running time of the program would have no practical end because of its
asymptotic behavior.

This ties in with a counter to one of the misnomers of using a total
language, which is that one may believe that it can not entail
long-running processes, servers, simulations, or other such programs.
The answer to this is that it is sufficient to have termination be
contingent upon external input, time, or a certain number of steps.
These are all more or less admissible for a flexible termination
checking step. Like the above example, the question of whether a program
terminates is generally interpreted in a \emph{syntactic} and
\emph{structural} sense, and becomes a lot more involved when
considering ``external'' effects.

Whatever paradigm is chosen, it is important to remember that legible
source code is paramount. This does not equate with terseness. It is
going to be a lot more helpful to have source code that some would
consider to be verbose, especially in the early stages where the
labeling problem remains unsolved.

\hypertarget{program-complexity}{%
\subsection{3.6 Program Complexity}\label{program-complexity}}

Another major concern is how to reduce program complexity. While related
to program search in the previous subsection, it has several
distinctions. Evaluation of fitness during program search does not
necessarily correspond with source code legibility. And that is what
separates it so cleanly from the research into iterative or sequential
program search. While a shorter program is preferable in many cases,
that has to be reconciled with the requirement for human understanding.

There are several dimensions to program complexity, and each of them
will be discussed. They are listed below by order of preference:

\begin{enumerate}
\def\labelenumi{\arabic{enumi}.}
\item
  The impact of program complexity on human understanding and program
  legibility.
\item
  The relationship between program complexity and model effectiveness,
  including fitness and length of descriptions in an information
  theoretic interpretation.
\item
  Program complexity in terms of runtime performance.
\end{enumerate}

The space of models is quite vast in machine learning, and it is often
the case that there are multiple models to choose from for a given set
of training data. The question of effectiveness then has to be
determined by a comparison with the other competing priorities of
complexity. And it should be noted that, apart from the first rule of
legibility, the other priorities should be taken not as a rigid
ordering, but as a set of guidelines; it is the specific use case of
each TML model that must ultimately determine the ordering of these
priorities.

\hypertarget{complexity-and-legibility}{%
\subsubsection{3.6.1 Complexity and
Legibility}\label{complexity-and-legibility}}

To maximize our understanding, we should first think about how we
develop high quality source code and then incorporate those ideas into
the TML algorithm. This could begin with an analysis of the structure of
programs, which should include function, block, statement, and
expressions each as independent levels of organization. The use of
multiple files or ``translation units'' is also a major consideration,
especially with regard to how it resolves dependencies, both internally
and externally, in the source model. A related question is whether the
system should involve the use of incremental compilation and linking. It
may be simpler for early TML systems to just focus on standalone
programs with a single translation unit.

Will we allow the TML system to optimize its own style or should it be
restricted? Care must be taken with this, as any and all limitations on
the expression of programs by the TML system could significantly impact
program search and model quality. This is especially difficult because
of the competing priorities to reduce program complexity. It may require
a lot of experimentation. Worse yet is that these checks and balances
may not be transferable between implementations; a set of priorities
that works for one TML system may not be beneficial for another.

Another aspect to program legibility is the question of which algorithms
and data structures we allow it to use to construct its model. Consider
this as a set of \emph{programming patterns}. Do we open up the search
space to include exploration of new patterns or should we provide this
set based on standard programming practice?

\hypertarget{complexity-and-model-fitness}{%
\subsubsection{3.6.2 Complexity and Model
Fitness}\label{complexity-and-model-fitness}}

All things being equal, descriptive complexity should be minimized, but
not at cost to legibility or runtime performance.

Operating over source code presents challenges for the process of model
search if it performs that search in a representation that is different
from the target languages of the TML model. While this additional layer
of abstraction is temporarily admissible, it brings with it the
additional requirement that the mapping it uses is \emph{provably}
invertible. This is a consequence of ensuring TML is verifiable.

It is tempting to think that the shortest program or subroutine is the
best, but this will not always be the case. The target architecture in
which the model will be deployed must be considered. Cache sizes, branch
prediction, memory constraints, and other factors play a role in the
balance between program complexity, size, and speed.

How, though, do we even approach the issue of balancing model fitness
with the complexity of its description? And how can we find or create an
algorithm that does this while giving us the ability to understand what
it produces?

One might begin by thinking that the TML algorithm must find a model
before it can explain it to us. But the way in which it constructs the
model will necessarily curtail the space of models in which it conducts
its search. And the optimization, measures, and metrics used during
search may be antagonistic to the simplest and most easily understood
programs. It has to know, in advance, what we consider to be legibile,
and that has to be used in tandem with any possible notion of fitness
for the model. They are inseparable.

So, the process of model search and program generation must be one in
the same, or at least highly interdependent. One does not, and can not,
come before the other. This is why a purely reductive approach using
data compression, information theory, and algorthmic complexity theory
could be extremely misleading to a researcher who is trying to design a
TML algorithm. Care must be taken to keep these measures in mind, but
not to be bound to them in a way that defeats their purpose.

\hypertarget{complexity-and-performance}{%
\subsubsection{3.6.3 Complexity and
Performance}\label{complexity-and-performance}}

One simple example of a major runtime impact is the calling conventions
that are used in the target language. For example, the System V AMD64
ABI provides that the first six arguments are passed in registers
\autocite{lu2018system}.

This places a premium on the complexity of the function signatures
themselves, tending towards six or less parameters on most consumer
64-bit hardware. On other target platforms, the calling conventions may
be completely different, and this could have a dramatic effect on not
only runtime performance, but the descriptive complexity of the model.
This would be the result of changes to the structure of function
signatures and the use of auxilary data structures to reduce register
pressure, just as a human programmer would do if faced with similar
constraints.

Earlier, in the Introduction section, the possibility to have both
legibility and efficacy without sacrificing either was mentioned. How
might that be achieved?

One way is that we utilize multiple modes. The default mode of the TML
system would be to produce programs that are less efficient, but easier
for us to understand. Then, once we understand the relevant parts of the
source model, we could instruct it to conduct a (super-)optimization
pass.

The TML would perform a translation between the original program and its
more efficient and effective representation, which we may understand
\emph{substantially} less. Nothing would be lost, however, as it would
be required to establish a rigorous mapping between the unoptimized
version and the optimized one. This would be done by using the so-called
\emph{de Bruijn criterion}, which would see it emit a full trace of
``proof objects'' in a format that could be externally verified with a
simple proof checker that we verify by hand and implicitly trust
\autocite{geuvers2009proof}. This would, in effect, give us the best of
both worlds.

\hypertarget{implementation}{%
\section{4 Implementation}\label{implementation}}

How might transparent machine learning be implemented? The purpose of
this paper is to begin an investigation to answer that very question.
All that can be provided right now is a list of suggested directions we
might take. It is expected that a great deal of research will be
required to find even a modest implementation of TML.

\hypertarget{early-stages}{%
\subsection{4.1 Early Stages}\label{early-stages}}

The first recommended direction would be to develop a flexible grammar
engine based on term-rewriting with capture-avoiding substitution. This
would need to be extremely efficient. A type system could be developed
within it. Dependent types are not considered a requirement for an
initial proof-of-concept.

It is not currently known whether or not it would be better to base such
a system on the simply-typed Lambda calculus or if it is best to have
the ability to arbitrarily specify formal grammars in a more open-ended
fashion. That is an open question likely to draw strong opinions from
various researchers.

After the grammar engine, the next step might be to see how this could
be used for program search and program generation. Of particular
importance is the need to confine the search space by ensuring that only
the correct forms for the language are generated. This sounds simple in
theory, but it is confounded by context in practice. This is because
there is a distinction between the syntax of a programming language and
its semantics.

One possible aid to this may come from something called the
Morphological Approach \autocite{zwicky1967morphological}. This,
however, is more of a way of thinking about the problem space than a
specific engineering solution. The usefulness of that approach is in
using it to find the internal consistencies of a particular grammar and
the corresponding semantics of its programs. This would then be used to
enhance the grammar engine so that it always remained correct with
respect to the particular programming language it was generating.

A more conventional approach would see the use of operational semantics
\autocites{plotkin1981structural}{schmidt1996abstract}{plotkin2004origins}{leroy2009coinductive},
but how that would be efficiently integrated with the proposed grammar
engine is unclear.

After those steps, the greatest initial obstacle will be the labeling
problem. This could be addressed early on with the use of labeled
training data. Research into ``semi-weak'' supervised learning
\autocites{ghadiyaram2019large}{yalniz2019billion} may also be used to
help with the process. Alternatively, there is the prospect of creating
a website where labels for training data could be crowdsourced with help
from the general public. Of course, that brings its own challenges, as
it would have to be filtered for spam, checked for accuracy, and
normalized against bias.

\hypertarget{research-directions}{%
\subsection{4.2 Research Directions}\label{research-directions}}

Below are some suggested research directions. While certainly not
exhaustive, it should provide more than enough for experimentation and
analysis in the early stages of developing transparent machine learning
algorithms. Each subsection will include a number of linked references
and will be listed, more or less, in ascending order of difficulty.

\hypertarget{intermediate-representations}{%
\subsubsection{4.2.1 Intermediate
Representations}\label{intermediate-representations}}

This approach could be helpful in providing a way to manipulate and
structure various programming languages and data formats. Regardless of
which one is chosen, it is highly likely that some form of intermediate
representation (IR) is going to be necessary to organize the structure
of the source models for TML.

The s-expressions \autocite{mccarthy1959recursive} found in the Lisp
programming language, and its many variants, are a significant potential
candidate for an IR, and would have the added benefit of being directly
accessible as part of an interactive programming environment. However,
this should not discourage the use of other formats, as the ability to
read and write interchange formats is ubiquitous in practice, and is
supported in some capacity by most general-purpose programming
languages.

Care, however, must be taken in the choice of the IR. There will be
overhead in its translation and processing. And, because of the burden
of proof requirements on translation, it will need to be representative
of the target languages without loss of information.

\hypertarget{interpreted-languages}{%
\subsubsection{4.2.2 Interpreted
Languages}\label{interpreted-languages}}

Early stage TML may be far simpler to explore with an interpreted
language, especially if it has a homoiconic structure or metacircular
\autocite{reynolds1972definitional} capability. Lisp dialects come to
mind, but the use of an interpreted language need not be restricted to
those forms. The intent is to provide universal support for
self-modification, regardless of hardware or operating system support.

It may be useful to create a universal or open framework for the
manipulation and representation of programming languages. The goal of
such a project would be to allow programs in one language to analyze and
transform the source code of itself or another program, regardless of
the programming language.

The main drawback to using interpreted languages, compared to native
compilation, is the penalty to runtime performance and memory
utilization. While memory use could be reduced, the cost of
interpretation is unavoidable. Anything other than the native
instruction set is going to incur a penalty. One technique to mitigate
this is the construction of threaded interpreters
\autocite{bell1973threaded}, which essentially use a computed form of
\texttt{goto} when evaluating instructions. Both GCC and LLVM have
extensions that support taking the address of a label, which can be used
for this purpose.

\hypertarget{bit-level-languages}{%
\subsubsection{4.2.3 Bit-Level Languages}\label{bit-level-languages}}

A bit-level language should not be confused with the notion of
interpretable byte code, though they are highly related. The main
difference is that bit-level languages are programming languages with an
ultra-compact representation. They are typically used to study
information theoretic and algorithmic complexity properties of program
descriptions. Notable examples include binary representations for the
lambda calculus \autocites{tromp2007binary}{grygiel2015counting}.

The parsimony of these languages could be of benefit to
super-optimization and model search, but at cost of having to translate
them into representations that we can understand. They are also arguably
less efficient than equivalent byte code due to the need to unpack bits.
On x86, for example, byte-aligned accesses for opcodes would allow
directly indexing an instruction table for the purposes of implementing
a threaded interpreter. To do this in a bit-level language the program
would have to perform several operations before an array index or table
offset could be calculated.

\hypertarget{just-in-time-compilation}{%
\subsubsection{4.2.4 Just-In-Time
Compilation}\label{just-in-time-compilation}}

The primary benefit of JIT compilation is that it combines the benefits
of (self-)interpretation with efficiency. This would be especially
useful if a TML system needed to perform real-time updates during
deployment. Generally speaking, the faster a TML system runs, the more
effective it will be at approaching optimal solutions.

Speed, however, is not the only consideration. Work towards the
acceleration of program search and program generation is ultimately
going to be a losing battle. While early efforts should be made to
improve the operational efficiency of TML systems, we must not lose
sight of the fact that it is the \emph{asymptotic behavior} of these
system that matters; for \(n\) iterations in program search,
O\((\log n)\) is vastly superior to O\((n)\) when \(n\) is large. And
the (micro-)optimization afforded by faster interpreters and program
execution only represents a reduction of \(k\) in O\((n + k)\) for the
same \(n\).

It should be seen as a measure of ``intelligence'' that one TML system
requires fewer \(n\) than another to obtain a source model. This would
remain true even if the models it produced were less effective. Even
more so if the reduction in iterations was significant. Such a TML
system would need to be studied in isolation to ascertain how it was
able to move through the program search space in such a way.

This also hints at the possibility of using TML to improve itself. This
is related to the field of meta-optimization
\autocites{stephenson2003meta}{branke2012meta}{neumuller2011parameter}{karpenko2011meta}{krus2013performance}.

\hypertarget{metamorphic-code}{%
\subsubsection{4.2.5 Metamorphic Code}\label{metamorphic-code}}

Metamorphic code is a program that is capable of rewriting itself in a
potentially different representation
\autocites{szor2001hunting}{li2011mechanisms}{bist2014detection}.

This would make a TML system capable of recognizing, disassembling, and
rewriting itself in pure machine code. The target language for the
source code model becomes the instruction-set architecture of the CPU.

At first glance, it would seem that having such a low-level
representation would defeat the purpose of legibility, but this is not
necessarily the case. An IR could be used, with a provable inverse, that
carried the additional information about the machine code so that the
representations could be presented side-by-side, with no loss of
fidelity.

A metamorphic program shares some overlap with JIT compilation in the
sense that it combines runtime efficiency with the ability to update
itself continuously. However, a distinction needs to be made between
online and offline updates. An online update is what a JIT performs for
a language with some form self-interpreting capability, such as
\texttt{eval()}. By contrast, an offline update is where a metamorphic
program modifies its own executable in the file system. This is not an
activity that a JIT would traditionally be expected to perform.

A standard program could be made to incorporate self-updates by loading
and unloading shared libraries while it is running, but this is no where
near the capability of a metamorphic program.

The most important consequence of a metamorphic engine is that it can
manipulate \emph{other} programs in machine code form. This capability,
when combined with TML, would give it the ability to modify compiled
programs, regardless of the language they were originally written in.
And, if the labeling problem were in its late stages of being solved, it
would prove to be a most effective disassembly and reverse engineering
tool. In its full form, TML could be used to forcibly ``open source''
executables. This would have a significant impact on cybersecurity.

\hypertarget{simulated-annealing}{%
\subsubsection{4.2.6 Simulated Annealing}\label{simulated-annealing}}

Simulated annealing is a powerful metaheuristic for global optimization
\autocites{kirkpatrick1983optimization}{vcerny1985thermodynamical}{moscato1990stochastic}{granville1994simulated}.
It may be especially applicable to the development of TML systems, as
program generation is discrete and covers a large search space. Parallel
simulated annealing, as discussed in
\autocites{aarts1988simulated}{greening1990parallel}{witte1991parallel}{pardalos1995parallel}{ram1996parallel},
may be of special interest to accelerating TML program search.

\hypertarget{accelerated-processing}{%
\subsubsection{4.2.7 Accelerated
Processing}\label{accelerated-processing}}

One of the greatest catalysts for modern ML is likely the fact that
these algorithms are so readily parallelized. This gives them the
opportunity to exploit the tremendous speedup from operating over
general-purpose graphics processing (GPGPU) architectures. That is a
very important property, as not all algorithms are created equally in
this regard.

The question must be asked: Does this same advantage apply to
transparent machine learning? Can the grammar engine at the core of TML,
or something equivalent to it, be implemented in a compute shader? And,
in general, can program search and program generation be done in a
massively parallel fashion?

If L-systems are any indication, then the answer to this question might
be in the affirmative, as demonstrated in
\autocites{mech2003generating}{lipp2010parallel}. An L-system, or
Lindenmayer system, is a type of rewrite system that can be used to
produce complex, fractal-like images and animations for computer
graphics \autocite{mccormack1993interactive}. They are naturally
parallelizable, however, and may be insufficient in power to iterate
over a general program space. Regardless, studying how these algorithms
are used in graphics hardware could result in a transfer of knowledge to
the TML domain.

If it turns out to be too difficult to write a grammar engine of
sufficient complexity in shader form, then the next best option is to
use a CPU-based method to target graphics accelerated hardware. There
has been some valuable work done using rewrite systems in this way to
generate OpenCL and compute shaders for GPGPU purposes
\autocites{steuwer2015generating}{steuwer2016matrix}. There is also
PyCUDA and PyOpenCL \autocite{klockner2012pycuda}. And of relevance may
be automatic termination analysis for GPU kernels
\autocite{ketema2014automatic}.

The problem with generating code for accelerated hardware is that it
limits updates and throughput. It also significantly complicates
verification. The ideal realization would be a ``pure'' GPU TML engine.
In principle, such a system would run entirely within the graphics
pipeline, and it would not require recompilation, except in the most
extreme cases where the TML engine was being changed itself.

One approach to a pure GPU solution may be to find mappings between
rewrite operations and data parallel instructions in the shader
language.

Another approach, though highly speculative, would be to emulate a
distributed virtual machine entirely on the GPU. The TML system would
then be executed in that environment. It should be noted that this is
distinct from the research on making GPU-accelerated hardware accessible
to a virtual machine instance. By contrast, this approach would
implement the TML system within the virtual instruction set being
emulated by one or more shaders.

Even if the above methods succeed, it will still present challenges for
systems level access and general input-output. In all likelihood, a
hyrbid CPU-GPU approach will be required, and this will, unfortunately,
complicate the source models generated by TML; it implies multiple
target languages and data formats, along with the additional overhead
that brings. Optimizing for both the CPU and GPU portion of a TML source
model could be an extremely complex challenge. It may be far simpler in
the early stages to create CPU-based TML models and then treat the GPU
target as an optimization pass.

\hypertarget{machine-consciousness}{%
\subsubsection{4.2.8 Machine
Consciousness}\label{machine-consciousness}}

The largest obstacle to full transparent machine learning is going to be
the labeling problem.

While various methods for labeling and classification are likely to be
developed, the author believes that they will reach a hard limit without
a fundamental advance in our understanding of artificial intelligence.

A basic component of that limit is related to something called the
\emph{grounding problem}, which is concerned with how the symbols in any
mental or cognitive model could be given meaning
\autocite{harnad1990symbol}. Instead of symbols, however, in the context
of TML, we have the inhabitants, elements, or objects of one or more
data types. And instead of a connectionist architecture, we are
constrained to source code we can understand in one or more programming
languages.

A full solution to the labeling problem must include an answer to the
grounding problem. There will be many other challenges for legibility,
but this is the most important first step towards a comprehensive
solution.

One way of approaching this problem would be to mimic consciousness. In
particular, the phenomenology of our subjective experience. This would
mean creating one or more streams of information for sensory perception
and then combining them into a unified whole. These could then be used
to build episodic memories that can be referred to by the learning
system, effectively grounding its representations in a form of
\emph{artificial sentience}.

It may be helpful to think of the primitives for this sensory data as
\emph{fragments of experience}, not unlike the texture fragments
manipulated in graphics shaders. In fact, that may be a most useful
analogy for thinking about how such data would be combined and processed
in a real implementation, especially given the distributed and parallel
nature of our biology.

There is another aspect to this future direction that should be
discussed as well. It is the importance of time, and how it is
consistently overlooked in how we currently design, build, and think
about AI.

The author strongly believes that any possible approach to resolving
machine consciousness must be done with respect to \emph{time}. It is
not sufficient to merely involve agents in some perception-feedback
loop. It has to be much more fundamental than that; the very
\emph{meaning} of the fragments of experience must co-vary with the
relative timeframe in which they are interpreted.

While this may all seem philosophical, it has an important connection to
any technical realization of machine consciousness: it demands the use
of a \emph{hard real-time system}. Whatever perceptual processing is
done, it will need to be done under strict deadlines, otherwise it will
change the interpretation and meaning of the experience. This is
contrasted to a soft real-time system, where missing some deadlines is
not considered to be a failure. See \autocite{shin1994real} for more
information on real-time computing.

The reason artificial sentience must be developed under hard real-time
constraints is because its meaning co-varies with time. Consider an
audio recording of someone saying ``one, two, three''. Imagine playing
this back at one-quarter speed. While we could eventually make out what
it is saying, and recover the number sequence, that sequence would not
represent the totality of the information about the recording. The
\emph{experience} of listening to it comes first, and our
\emph{interpretation} of that experience is secondary.

All of this would make machine consciousness, and the transparent
machine learning systems built over it, akin to a simulation. This
approach would form the basis for a \emph{cognitive architecture}, which
opens up the possibility of building a more general-purpose framework
for studying artificial intelligence as subjects of experience. And
there is even more to discuss on this topic\footnote{See also: ``Machine
  Consciousness'', AI Security (Juliano, 2016).}, but, unfortunately, it
would be far beyond the scope of this article. This was presented only
as a potential research direction towards solving the labeling problem.

\hypertarget{theoretical-limits}{%
\section{5 Theoretical Limits}\label{theoretical-limits}}

Will human understanding in TML scale as the overall complexity of its
models increase? Is there a threshold of program complexity beyond which
transparent ML \emph{necessarily} degenerates to opaque ML?

A distinction must be made between program legibility and the ease in
which it can be understood. This is not to contradict the standard test
of reasonable competence set previously in the definition of TML. It is
meant to facilitate discussion about the theoretical limitations of TML
source models at the boundary of maximum model complexity and complete
human understanding.

Like natural language, source code can entail complex logic,
mathematics, algorithms, and patterns that one can recognize as correct
statements of the language, but would otherwise require years of study
to fully understand.

It is useful then to imagine a TML system that, while fulfilling its
stated obligations, reaches a point where it starts to produce models
that are so sophisticated that its burden of satisfying human legibility
begins to weaken its expressive power.

To help discuss this, two complementary definitions from Yampolskiy
\autocite{yampolskiy2019unexplainability} on the foundations of AI will
be considered:

\begin{enumerate}
\def\labelenumi{\arabic{enumi}.}
\item
  \textbf{Unexplainability:} For certain decisions made by an
  intelligent system there will be no explanation that is both 100\%
  accurate and comprehensible to humans.
\item
  \textbf{Incomprehensibility:} Certain decisions made by an intelligent
  system will have a 100\% accurate explanation for which no human can
  completely understand.
\end{enumerate}

The first step in addressing these claims is that we must consider the
distinction between opaque ML and transparent ML. In TML, the
explanation is \emph{inseparable} from the model, because it is the
model. And that model may also include the description of the TML system
itself.

A separate description that explains an opaque ML model may be of low
accuracy and it will not affect the deployment of that model. This is
not the case with TML, as the description and the model are not just
equivalent to each other, they are the same object. So, any description
of a TML model is of maximum accuracy by definition.

This simplifies at least one part of the analysis: there are no
inaccurate descriptions of models while remaining within the framework
of TML. Every TML system must produce models that are valid programs.
This is because every TML model has a source code component that must be
recognized by a compiler or interpreter for the target languages, which,
by the TML requirements, must include at least a syntax and type check.
The TML system then performs additional checks during program search,
program generation, and runtime evaluation based on a balance of
priorities for program complexity and model fitness.

While the author acknowledges the possibility of there being written
documentation, scientific theories, or mathematical theorems resulting
from knowledge gained from a TML model, these must be seen as
supplemental. The definition given in (1) for unexplainability entails a
spectrum of comprehensibility that co-varies with the accuracy of
descriptions for machine learning models. As the description for a TML
model is inseparable and exact by definition, this makes accuracy a
constant factor, leaving only comprehensibility. Under such an
interpretation, this effectively reduces claim (1) to claim (2), which
should not be surprising, as they are complementary to begin with. Under
TML they are simply equivalent.

This is a necessary first step towards addressing incomprehensibility.
The rest of the subsections will proceed based on this clarification.

\hypertarget{legibility}{%
\subsection{5.1 Legibility}\label{legibility}}

In TML, program legibility is intended to be synonymous with human
comprehensibility, which was given in terms of reasonable standards of
skill.

The system has to take this legibility requirement into account when
performing its search. This means that not only is a model's description
or explanation inseparable from its construction, the effect of
legibility on program complexity is necessarily a part of the search for
the model. The model does not come first; legibility is not evaluated,
but likely a fundamental part of the generative process. That is another
way in which it is so distinct from opaque ML.

Legibility places a constraint on the space of all possible programs
that TML can generate, limiting it to only those programs that
necessarily satisfy user-defined legibility. That will exclude otherwise
valid programs from the search space, and is highly relevant when
interpreting claim (2) of incomprehensibility.

We will not get incomprehensible models if a TML system is operating
correctly. This can be seen as (2) placing a maximum upper-bound on TML
model \emph{effectiveness} in the limit of program legibility. This
would be caused by a TML system not being able to adequately express
programs because they do not combine in such a way as to satisfy
expected results from the input training data.

It was suggested in the section on program complexity and model fitness
that we utilize multiple representations, provided that we have a
provable inverse. The problem with this is that it shifts the burden of
legibility to finding highly expressive instruction sets and programming
languages with inverses that produce models we can reverse engineer into
legible data structures, algorithms, and programming patterns. While
probably easier than interpreting deep neural networks
\autocite{montavon2018methods}, it would still be an oblique approach to
the problem of legibility.

One might try to have a TML system optimize its ability to make legibile
programs in an effort to strengthen its expressive power. That would be
difficult, as TML relies upon a user-defined notion of what it must
consider to be legible, which was previously referred to as a set of
programming patterns. It has no general criterion for what any human
would consider more or less preferable in terms of legibility, and this
limitation would still apply even if general intelligence were to be
involved. This can be observed in the fact that even different human
programmers have widely varying standards program quality.

\hypertarget{groups}{%
\subsection{5.2 Groups}\label{groups}}

Consider the billions of lines of code in all of the actively maintained
and publicly available free and open source software projects that
currently exist. These are developed and maintained by one or more human
programmers, some with the aid of automation in varying capacities. Each
of these individuals have their specific talents, knowledge, and areas
of expertise.

It is inconceivable that any \emph{one} of us could ever \emph{fully}
understand \emph{all} of these projects in totality. This is direct
evidence for claim (2) in the more general sense. Even though the
individual parts of this collective work are in fact understood by those
working on it, the sheer volume of all these projects in combination
would \emph{necessarily} exceed human ability to follow.

That evidence, however, also contains a strong counter-claim. By adding
one more human we have been able to scale with the growing complexity in
software and hardware. We specialize and work in teams. Even if we are
not directly communicating or collaborating, our combined efforts are a
form of collective human understanding. This is not an implicit argument
for emergence, but a statement of fact that we do overcome immense
complexity through group effort.

These same arguments apply to the totality of human knowledge in general,
but the specific case of source code projects was used because it
exemplifies the kind of complexity we might anticipate from TML models
of extraordinary sophistication.

Consider a TML model that has grown in complexity to the extent that it
is in the hundreds of billions of lines of code. If each of the parts of
the model satisfied the legibility requirements of the TML definition
then there is no reason, in principle, that we could not scale to the
challenge by adding one more human to the problem, and repeating that
step until we have collectively understood it well enough.

Unfortunately, if it turns out to be the case that a full-spectrum
general AI, or superintelligence, must necessarily involve a model of
such size, then we may find ourselves unable to respond quickly enough
to anticipate it, despite being able to comprehend its description
through a concerted group effort.

In such a case, the author suggests that we construct a \emph{trusted
sequence} of artificial intelligence where each element has a strictly
increasing ability to comprehend its successor better than its
predecessor. This would still incur delays in prediction and analysis,
but at possibly shorter timescales than that of a comparative group
effort made by humans.

\hypertarget{data}{%
\subsection{5.3 Data}\label{data}}

It was anticipated that there might be a need for data to accompany the
programs representing complex TML models. It is for this reason that the
ability to target data formats and markup languages was explicitly
included in the definition.

But how does that relate to incomprehensibility?

Consider a TML model that was equal or better than us at recognizing
human faces. It is reasonable to expect that it would have both a data
and source code component. This would mean that the TML system would
need to target at least two distinct languages, with one of them being a
data or markup language and the other being a programming language.

The data part of the TML model might be a listing of the numeric values
for the symmetries and structures it had found for the faces in the
training data. That might include composite models or topological
information in its own unique encoding. The author argues that this
would be \emph{analyzable} by data scientists, even if they had to first
find a more meaningful representation or data visualization technique.

On the source code side, the model would use that data in queries or
lookups to simulate, project, or manipulate incoming image data through
the use of one or more algorithms and data structures.

While the goal of TML is to produce legible programs, this does not
necessarily apply to the \emph{content} of the targeted \emph{data}
languages. In other words, while the data languages used must be legible
to us, and their statements well-formed, this does not mean that the
data will be anything other than just a \emph{direct representation} of
the information that is vital to the operation of the source model.

It is the design intent of TML that the data structures, algorithms, and
programming patterns ultimately determine legibility, even though what
is legible is user-defined. While data will play a role in the
construction of the particular models, it should not be taken as a limit
on model effectiveness, legibility, or complexity.

On the other hand, that design intent must not be used as an excuse to
make space versus time trade-offs that reduce program complexity while
sacrificing overall clarity. This could be taken as a qualification on
the legibility condition. The purpose of including data language targets
was to make a practical distinction between programs and data, even
though they are both a form of information.

A TML system should minimize the data portion of its model while
simultaneously maximizing program legibility, without sacrificing the
other competing priorities on complexity and model fitness.

Following the example given, it is reasonable to accept that the
geometric symmetries in millions of human faces might be represented
best in a numeric interpretation, even though an algorithm could be
presented that generates it with a much shorter description. The TML
model that produces that data would, in fact, be one example of such an
algorithm, for the simple fact of having generated it. That, however,
does not mean that the TML model and its \emph{training} data should be
admitted as an explanation of human facial recognition.

What we seek in practice is a generalization of that training data in
the form of a model, without the need for its specific examples again in
the future. The data portion should be taken as the information portion
of the TML model that could not be effectively represented in source
code form, or was otherwise classified as the input to the source code
portion of the model.

The TML approach makes an intentional distinction between data and
source code for this purpose. It reduces the length of programs and
enables alternative constructions to accelerate them.

It is the data side of TML that can be used to spare it from claim (2),
which would permit the source code to remain legibile as overall model
complexity increases.

A data set may very well be incomprehensible to us, but would not
prevent us from analyzing, visualizing, and studying it. There is
nothing, in principle, that prevents such data from being the useful
input to algorithms we \emph{do} understand. Additionally, the quantity
of the data need not affect our understanding of the algorithms that use
it, but it could add time to an analysis of the model as a form of
knowledge.

\hypertarget{output}{%
\subsection{5.4 Output}\label{output}}

As TML models are just computer programs, the study of algorithms and
computation directly apply to every model produced by any possible TML
system. The following analysis will focus on the output of these models
by approaching it from the perspective of algorithms and computation. In
this context, and unless noted otherwise, the word ``model'' should be
considered as a TML model, which makes it synonymous with the word
``program''. Where it makes sense, these terms will be used
interchangeably for the remainder of this section.

It is possible to understand the description of a program while not
being able to easily predict its runtime behavior
\autocite{zwirn2013unpredictability}. Such runtime behavior can further
be divided into the internal state of the program and its output, with
the latter being most relevant to this analysis.

This brings up an important point about program legibility in the
definition of TML. The goal of this approach is to produce \emph{models}
that we understand, and every model that is generated by TML should
produce behavior that is useful, according to some user-defined criteria
for model fitness. The definition of TML does not specify any
requirement on the predictability or comprehensibility of the runtime
behavior of these programs, and there are several reasons for this.

The comprehensibility of model output is not contingent upon our ability
to predict that output in advance. Safety notwithstanding, we can still
benefit from models we can not predict, just as we do with natural
intelligence; we do not need to predict the wording of the next
scientific paper for that result to be both comprehensible and useful to
us.

It may even be the case that the unpredictability of model output is
impossible to avoid \autocite{yampolskiy2019unpredictability} or is an otherwise \emph{necessary condition} for the most effective models.
Instead of a limitation, however, the author argues that such a property
of model output could be interpreted as novelty or creativity, which
would be a highly desirable outcome. This should be possible, at least
in principle, as TML has the ability to sample from the space of all
possible programs. But how would that enable such output
characteristics?

Consider a model with one or more nondeterministic algorithms
\autocite{floyd1967nondeterministic}, each of which are using external
input in a stream of unbounded processing. Now give it sufficient memory
and storage so that the program can refer to previous external inputs
and runtime states. Such a model would have the potential to generate an
unbounded amount of new information from an otherwise finite
\emph{initial} program description. And there is no reason to believe
that TML would not be able to produce such models in theory.

So, if unpredictability of output is unavoidable for the most effective
models, and even a desirable indicator of novelty, then that leaves only
the comprehensibility of that output to consider.

Suppose it is possible for full-spectrum general intelligence or
artificial superintelligence \autocite{bostrom2016superintelligence} to
be represented as programs. These kinds of AI would then form a subset
of the set of all possible programs. Now devise a TML system that is
instructed to include that subset within its search space. This would
lead to one of the following outcomes:

\begin{enumerate}
\def\labelenumi{\arabic{enumi}.}
\item
  These advanced models would not be found by any TML system, as the
  human legibility requirements would preclude it from expressing models
  at that level of sophistication.
\item
  It would find one or more models at that level of sophistication, and
  we would understand them, but they would produce output that we can
  not understand, even with a group effort.
\end{enumerate}

The author believes that the second scenario is the most likely outcome.
This represents a much worse result, however, as it strengthens the case
for a different kind of incomprehensibility. We would, in principle,
have the ability to understand some of the most sophisticated AI models
possible, but the price would be that its outputs would be
incomprehensible to us.

There are a couple of ways in which the output of such models might
become incomprehensible. It could be at a level of abstraction or
sophistication that is beyond human ability to follow. Even if we could
eventually find one or more people to specialize in the knowledge it
produced, it may take us too long, and, by then the model might have
produced something even more complex. This ties in with the next
concern, which is the rate at which these models could produce novel
output. If done at scale, it could preclude human understanding even under
a collective interpretation.

One can imagine humanity falling behind an exponential curve of
incomprehensibility from such models of intelligence. Even if our entire
population perfectly coordinated and specialized in the knowledge it was
amassing, we might never hope to keep up.

\hypertarget{closing}{%
\section{6 Closing}\label{closing}}

Transparent machine learning has been introduced as a possible
alternative direction in machine learning. It would give us the ability
to produce explicit AI that we can study, verify, and refine. This would
transform the way we integrate automation with our technology by
leveraging the existing hardware and software development processes that
are commonplace today.

More importantly, TML answers the question of how we might control AI,
both now and in the future. Having the source code to the implementation
is a common sense prerequisite, and one that TML would provide. This
would allow us to ensure that these systems behave exactly as we expect
and enable us to program them to exacting specifications, in the most
transparent and verifiable way possible. That includes codifying our
ethical norms and laws within these systems. While TML does not address
how we decide or translate our values into that form, it does ensure
that it will be an explicit part of the implementation. And it will be
present in a form that we can directly check.

The benefits of learning from TML source models can not be overstated.
This could be of aid to an early form of automated science, where
working theories are generated directly from data. Domain-specific
languages could be used to translate source code to and from the
language used by specific scientific disciplines. This may be one of the
most direct routes to constructing automated research assistants until
the discovery of general AI has been made. It only takes a willingness
to see source code as a form of knowledge.

To balance the benefits of TML, a warning and disclaimer must also be
given. As with any new technology, there is a significant potential for
misuse of TML. By definition, TML has the ability to read and write
source code. If combined with a metamorphic engine, this would also give
it the ability to read and write any program it can access, without the
need to execute it. This includes updatable hardware as well.

Metamorphic software with AI capability has to be treated as its own
class of malware. The very same creative tools that will advance TML
research can also be used to destroy. The upper limit on what this kind
of malware can do should be treated as equivalent to what a human
operator would be able to do with physical access to a compromised
system. While that full threat potential is not realizable now, it may
become reality later when TML becomes more sophisticated.

The suggested defensive action with TML is to utilize it to harden
individual programs and networks. An API could be developed that exposes
the network directly to the target language. The TML system could then
be instructed to make program search and generation with that API a part
of its model. This could, in principle, enable it to manipulate and
explore that network through the use of the API, finding vulnerabilities
and other flaws in security.

The true power of TML is that it can be used to search, iterate, and
generate sequences over arbitrary formal grammars. And it could be
employed for any problem domain that can be described in such a way. It
is distinct from conventional ML because TML could do this with little
or no training data whatsoever. This is because it exploits the formal
structure of the grammar. Consider the case where that grammar entails
genetic, chemical, or physical models. As long as a reasonable measure
of fitness could be developed, then TML could be used to investigate the
space of sequences for the respective target domain without it having to
necessarily be a computer program.

Lastly, the pursuit of full TML will likely converge with research
directions for general AI. These technologies would be complementary to
each other, especially in terms of trust and security. The labeling
problem will be the greatest challenge going forward, and it may require
research into machine consciousness to resolve. That, in turn, could
help unlock new ways of thinking about general AI, which could and
should be considered the ultimate objective of TML related research.

\printbibliography[title=7 References]

\end{document}